\documentclass[runningheads]{llncs}

\usepackage{graphicx}
\usepackage{amsfonts}
\usepackage{capt-of}
\newcommand{\revision}[1]{#1}
\usepackage[hidelinks]{hyperref}

\title{CI-JEPA: A Counterfactual Analysis of Latent Representations in Joint-Embedding Predictive Architectures for Self-Supervised Learning}
\titlerunning{CI-JEPA}
\author{Mintu Dutta\inst{1} \and Ritesh Vyas\inst{2} \and Mohendra Roy\inst{1}\thanks{Corresponding author: Mohendra Roy, Email: \email{mohendra.roy@ieee.org}.}}
\authorrunning{M. Dutta, R. Vyas, and M. Roy}
\hypersetup{
    pdftitle={CI-JEPA: A Counterfactual Analysis of Latent Representations in Joint-Embedding Predictive Architectures for Self-Supervised Learning},
    pdfauthor={Mintu Dutta, Ritesh Vyas, Mohendra Roy},
    pdfsubject={Counterfactual intervention-aware self-supervised visual representation learning},
    pdfkeywords={Self-supervised learning, I-JEPA, counterfactual interventions, explainable AI}
}
\institute{Department of Information and Communication Technology, Pandit Deendayal Energy University, Gandhinagar, Gujarat, India \and Department of Electrical and Electronics Engineering, Birla Institute of Technology \& Science, Pilani, Rajasthan, India}

\begin{document}
\maketitle
\begingroup
\makeatletter
\let\@footnotetext\H@@footnotetext
\makeatother
\renewcommand{\thefootnote}{}
\footnotetext{Accepted for the International Symposium on Smart Systems, Algorithms \& Applications 2026 (IS$^{3}$A$^{3}$ 2026), Tezpur Central University, India}
\endgroup
\begin{abstract}
Self-supervised visual representation learning learns useful features without manual annotations during representation training. The image-based joint-embedding predictive architecture (I-JEPA) predicts latent representations of masked image regions, but its objective does not explicitly model responses to specified visual interventions. We introduce CI-JEPA, a counterfactual intervention-aware extension that learns to predict the representation change $\Delta Z = Z_{\mathrm{CF}} - Z$ between an original image and a modified counterpart. We assess representation robustness through selective sensitivity: stronger responses to task-relevant semantic changes than to nuisance changes. Experiments on Flowers102 use flower-center occlusion as a candidate semantic intervention and background blur and tint as candidate nuisance interventions. With frozen-encoder linear probing, CI-JEPA achieves a best validation accuracy of 78.14\%, compared with 77.55\% for both the pretrained ViT-B/16 and the I-JEPA baseline, a gain of 0.59 percentage points. The reported mean $L_2$ representation changes are 4.42 for center occlusion, 3.48 for background tint, and 2.83 for background blur. This ordering is consistent with relative semantic selectivity for the evaluated interventions, rather than complete nuisance invariance. The accuracy comparison is complementary and does not establish improved robustness over the baselines. These controlled image modifications provide a framework for studying intervention-induced changes in JEPA representations; they do not establish causal feature discovery or robustness to all visual changes.

\end{abstract}

\section{Introduction}
Deep learning models can achieve high recognition accuracy, yet the information they use to make decisions can be difficult to identify, especially in self-supervised learning. For example, a model may correctly identify a flower using its visual characteristics, but it may also rely on the background or other incidental information. Self-supervised learning aims to learn useful visual features without labels during representation training. I-JEPA predicts the latent representations of masked image regions~\cite{assran2023ijepa}. Because it operates in representation space rather than pixel space, its objective encourages the prediction of higher-level semantic information. The architecture can be expressed as follows:

\begin{equation}
\mathbf{z}_C=f_{\theta}(\mathbf{x}_C),\qquad
\mathbf{z}_T=g_{\phi}(\mathbf{x}_T),\qquad
\hat{\mathbf{z}}_T=q_{\psi}(\mathbf{z}_C,\mathbf{m}_T).
\end{equation}

The input image is divided into a visible context region $\mathbf{x}_C$ and a target region $\mathbf{x}_T$. The context encoder $f_{\theta}(\cdot)$ maps the visible context into a latent representation $\mathbf{z}_C$, while the target encoder $g_{\phi}(\cdot)$ produces the corresponding target representation $\mathbf{z}_T$. The predictor $q_{\psi}(\cdot)$ uses the context representation together with the positional information $\mathbf{m}_T$ of the target region to predict its latent representation $\hat{\mathbf{z}}_T$. The learning objective is expressed as:

\begin{equation}
\mathcal{L}_{\mathrm{I\!-!JEPA}}
=
\frac{1}{|T|}
\sum_{i\in T}
\left\|
\frac{\hat{\mathbf{z}}_{T,i}}
{\|\hat{\mathbf{z}}_{T,i}\|_2}
-
\frac{\mathbf{z}_{T,i}}
{\|\mathbf{z}_{T,i}\|_2}
\right\|_2^2.
\end{equation}

The model is trained by minimizing the distance between the predicted target representation $\hat{\mathbf{z}}_{T,i}$ and the target representation $\mathbf{z}_{T,i}$ for each target patch $i \in T$. Both representations are $L_2$-normalized before computing the squared Euclidean distance. Consequently, the model learns to predict the target in representation space rather than reconstructing its pixel values.

Here, $T$ denotes the set of target patches, $|T|$ is the number of target patches, $\mathbf{z}_{T,i}$ is the target encoder's representation of patch $i$, $\hat{\mathbf{z}}_{T,i}$ is the predictor's estimate of that representation, and $\|\cdot\|_2$ denotes the $L_2$ norm. The fundamental learning objective is therefore:
\begin{equation}
    \fbox{$\hat{\mathbf{z}}_T \approx \mathbf{z}_T$},
\end{equation}
rather than
\begin{equation}
    \hat{\mathbf{x}}_T \approx \mathbf{x}_T.
\end{equation}
This is the key distinction between I-JEPA and pixel-level reconstruction methods.

The parameters $\phi$ of the target encoder are updated using an exponential moving average (EMA) of the context encoder parameters $\theta$. Here, $\tau \in (0,1)$ is the momentum coefficient. Unlike the context encoder and predictor, the target encoder is therefore not directly optimized through backpropagation of the prediction loss. Instead, it evolves as a slowly changing version of the context encoder, providing a stable target representation.

\begin{equation}
\phi\leftarrow\tau\phi+(1-\tau)\theta.
\end{equation}

\subsubsection{Motivation for the Counterfactual Block}
I-JEPA predicts target-region representations from visible context~\cite{assran2023ijepa}. This objective encourages semantic prediction, but does not explicitly supervise how a representation should respond to a specified visual intervention. Strong recognition accuracy alone therefore cannot establish whether the learned features reflect the object or incidental correlations. Evidence for this concern comes from Xiao et al.~\cite{xiao2020noise}, who showed that object-recognition models can classify images using backgrounds alone and can change their predictions when backgrounds are replaced while foreground objects are retained. These findings motivate testing background dependence; they do not establish that I-JEPA exhibits the same behavior on Flowers102.

Our central question is whether an encoder responds differently to changes in potentially discriminative object information and changes in nuisance background information. We use flower-center occlusion as a candidate semantic intervention and background blur and tint as candidate nuisance interventions. We hypothesize that the former produces a larger representation change than the latter. This distinction is an experimental assumption rather than a universal property of flowers: center information may vary in importance across classes, and intervention magnitude or editing artifacts may also influence the measured response.

The counterfactual block makes this question part of the learning objective. Paired original and modified images provide a target change, $\Delta Z$, while the intervention-conditioned predictor learns to estimate that change. Unlike simply including modified images as additional training samples, this objective explicitly relates an intervention to its effect in latent space. It also avoids assuming that every modification should leave the representation unchanged, since removing relevant object information may legitimately change it. Counterfactual visual explanations have used spatial replacements to examine changes in class predictions~\cite{goyal2019counterfactual}; here, we instead train a predictor of intervention-induced representation changes.

\revision{The primary motivation for using counterfactuals in our experiments is to assess the robustness of the learned representations. We define robustness as selective sensitivity: the representation should respond to changes in semantic information while remaining relatively stable under nuisance changes. Thus, greater sensitivity to semantic counterfactuals than to nuisance counterfactuals is the desired behavior under our evaluation criterion.}

The main contributions of this work are as follows: (1) we add an intervention-aware counterfactual branch to I-JEPA; (2) we generate several controlled counterfactual versions of flower images; (3) we measure how each intervention changes the learned representation; (4) we compare nuisance changes, such as background modifications, with meaningful changes, such as flower-center occlusion; and (5) \revision{we assess representation robustness through the relative sensitivity to semantic and nuisance counterfactuals.}

\section{Background}
There is a substantial body of research on model explainability~\cite{ribeiro2016trust,lundberg2017unified,sundararajan2017axiomatic,selvaraju2017gradcam}, as well as on causal and interventional representation learning~\cite{scholkopf2021toward,ahuja2022interventional}. Counterfactual learning~\cite{goyal2019counterfactual} examines how a model would respond if part of its input were changed while other information remained the same. In computer vision, this is done by creating a modified version of an image and comparing the model outputs for the original and modified images.

Several studies have generated counterfactual examples to improve model explanations and test robustness. For example, ``Counterfactual Visual Explanations''~\cite{goyal2019counterfactual} generated counterfactual explanations by replacing selected spatial regions with regions from a distractor image to change the predicted class. ``Generating Visual Explanations''~\cite{hendricks2016generating} generated class-discriminative natural-language explanations of visual recognition decisions. Causal representation learning methods have also used interventions to examine how changes in underlying factors affect learned features~\cite{li2024causal,yang2021causalvae}.

Our work is related to these approaches but has a different focus. We apply controlled visual interventions directly to an I-JEPA-based representation learning framework. Rather than using counterfactual images only to explain the final class label, we measure how each intervention changes the internal representation. We also compare two types of changes: nuisance interventions, such as background modifications, and semantic interventions, such as covering the center of the flower.

\section{Proposed Methodology}
\subsection{Overview}
\begin{figure}[htbp]
    \centering
    \includegraphics[width=0.9\textwidth]{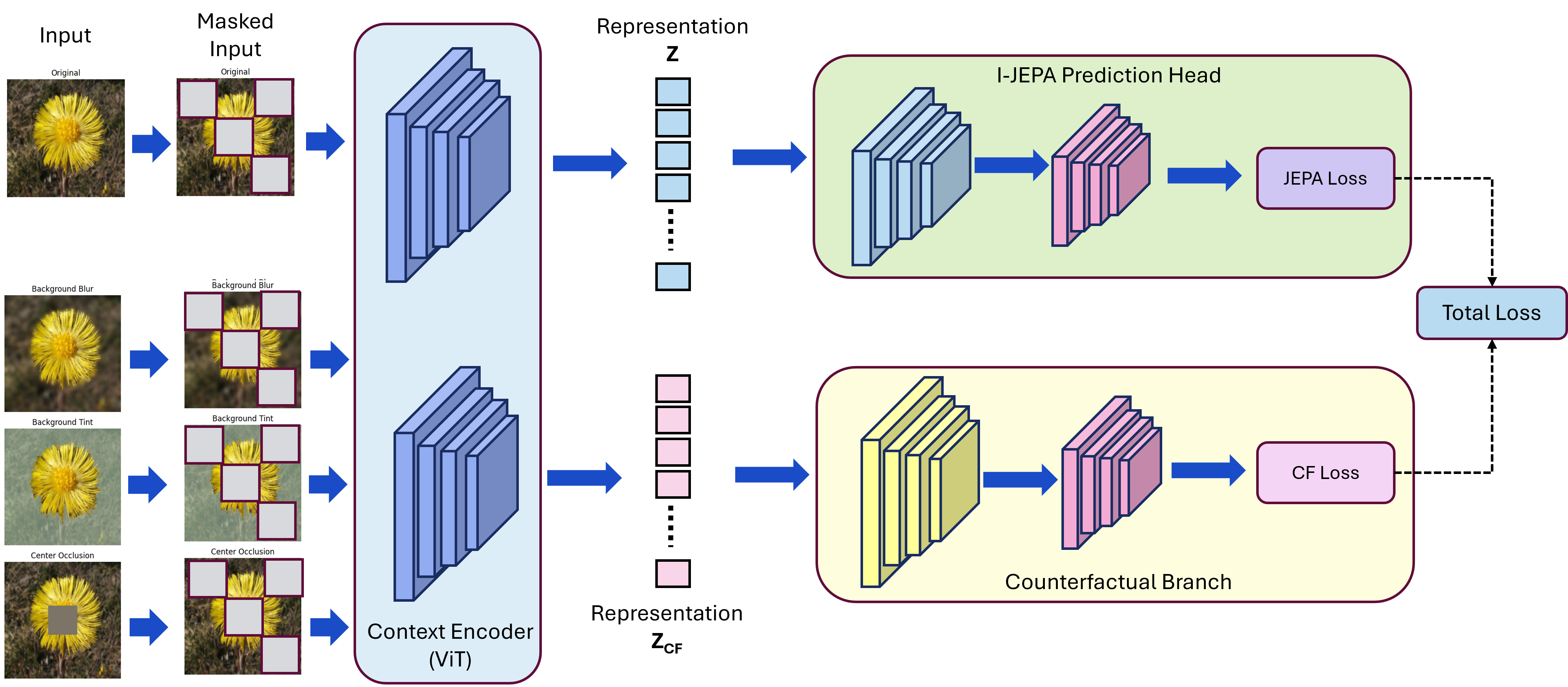}
    \caption{Proposed architecture for I-JEPA with Counterfactual Block.}
    \label{fig:counterfactual-ijepa-architecture}
\end{figure}

Our approach starts with the original I-JEPA architecture and adds a counterfactual branch. The original image is processed by an encoder to produce the representation $Z$. We then generate a counterfactual version of the image. The same encoder produces a new representation, $Z_{\mathrm{CF}}$, for the counterfactual image. The difference between these two representations indicates how strongly the intervention affected the model: $\Delta Z = Z_{\mathrm{CF}} - Z$. The overall architecture, illustrated in Figure~\ref{fig:counterfactual-ijepa-architecture}, integrates the original and counterfactual processing paths so that the predictor learns both masked-region representations and intervention-induced representation changes.

The counterfactual loss in Equation~\ref{eq:counterfactual-loss} supplies a complementary training signal: it measures how accurately the predictor models the change caused by each intervention, whereas the original objective measures masked-region prediction. Retaining both objectives is intended to preserve useful semantic features while making intervention responses predictable. The intervention embedding and binary mask provide conditioning information for this additional task; their contribution to robustness requires empirical evaluation rather than following automatically from the architecture.

\begin{figure}[htbp]
    \centering
    \includegraphics[width=0.9\textwidth]{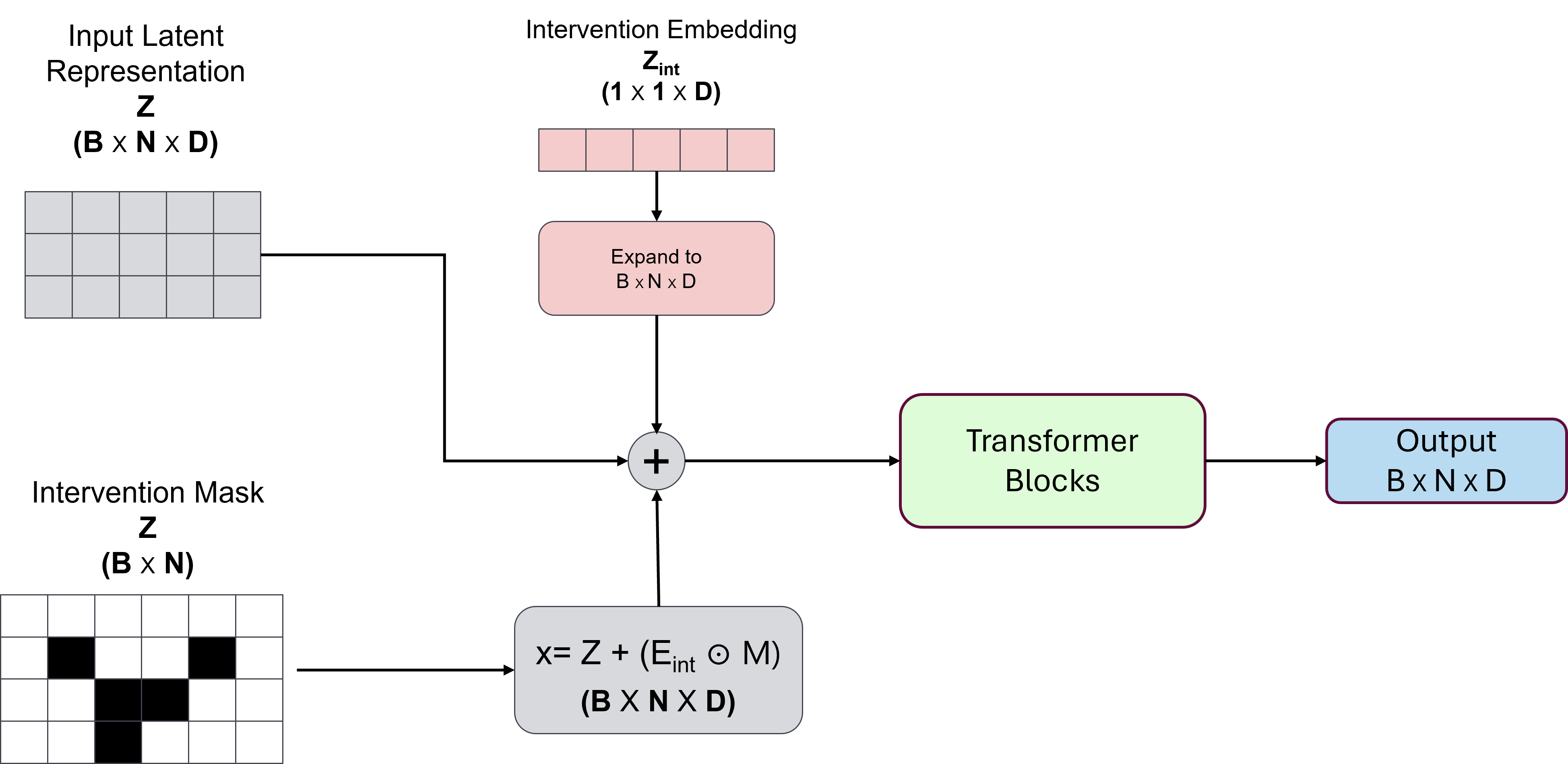}
    \caption{Architecture of the proposed counterfactual block.}
    \label{fig:counterfactual-branch}
\end{figure}

The counterfactual branch, shown in Figure~\ref{fig:counterfactual-branch}, is designed to estimate the counterfactual representation $Z_{\mathrm{CF}}$. It introduces a learnable intervention component $E_{\mathrm{int}}$ with an initial dimension of $1 \times 1 \times D$. For a training batch, this component is expanded to match the masked-token representation:
\begin{equation}
    E_{\mathrm{int}} \in \mathbb{R}^{1 \times 1 \times D}
    \quad \longrightarrow \quad
    \widetilde{E}_{\mathrm{int}} \in \mathbb{R}^{B \times N \times D},
    \label{eq:intervention-embedding-expansion}
\end{equation}
where $B$ is the batch size, $N$ is the number of masked tokens, and $D$ is the embedding dimension. The expanded intervention component conditions the predictor on the applied counterfactual transformation, enabling it to estimate $Z_{\mathrm{CF}}$ and the corresponding representation change. The intervention embedding $E_{\mathrm{int}}$ is randomly initialized and learned during training. The intervention mask is binary, where 0 indicates no intervention and 1 indicates an intervention.

For a given intervention $I$, the target representation change is defined as
\begin{equation}
    \Delta Z^{I} = Z_{\mathrm{CF}}^{I} - Z^{I},
    \label{eq:intervention-representation-change}
\end{equation}
where $Z^{I}$ and $Z_{\mathrm{CF}}^{I}$ denote the original and counterfactual representations, respectively. The counterfactual predictor produces an estimate $\widehat{\Delta Z}^{I}$ of this change. The loss associated with intervention $I$ is computed using the Smooth L1 loss:
\begin{equation}
    \mathcal{L}_{\mathrm{CF}}^{I}
    = \mathrm{SmoothL1}\!\left(\widehat{\Delta Z}^{I},\Delta Z^{I}\right).
    \label{eq:per-intervention-cf-loss}
\end{equation}
For a set of $K$ interventions, the final counterfactual loss is the mean of the individual intervention losses:
\begin{equation}
    \mathcal{L}_{\mathrm{CF}}
    = \frac{1}{K}\sum_{I=1}^{K}\mathcal{L}_{\mathrm{CF}}^{I}.
    \label{eq:counterfactual-loss}
\end{equation}

\subsection{Experimental Setup}
\subsubsection{Dataset}
For this experiment, we used the Flowers102 dataset~\cite{nilsback2008automated} (maintained by the University of Oxford), which contains 8,189 images. We used the official split of the dataset: 1,020 training images, 1,020 validation images (10 images per class), and the remaining 6,149 images for testing. A sample from the dataset is shown in Figure~\ref{fig:dataset-sample}.
\subsubsection{Counterfactual Interventions}
We designed two kinds of interventions: nuisance interventions (background tinting, blurring, etc.) and semantic interventions (flower-center removal). A sample of the counterfactual images is shown in Figure~\ref{fig:cf-sample}.

\begin{center}
\begin{minipage}[t]{0.48\textwidth}
    \centering
    \includegraphics[width=\linewidth]{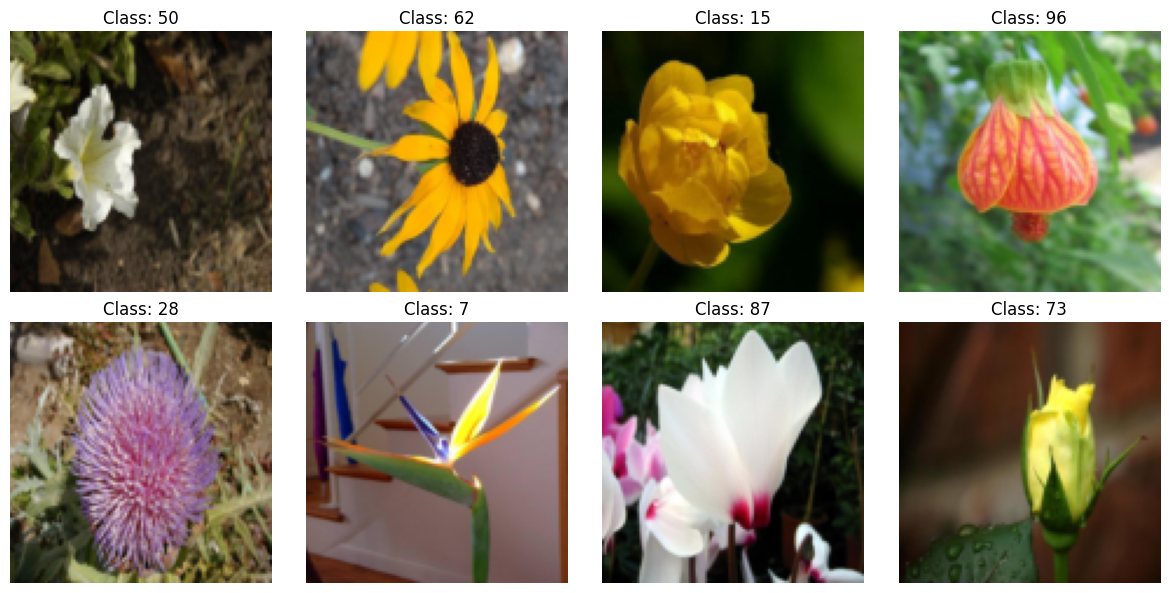}
    \captionof{figure}{Sample of the dataset.}
    \label{fig:dataset-sample}
\end{minipage}
\hfill
\begin{minipage}[t]{0.48\textwidth}
    \centering
    \includegraphics[width=\linewidth]{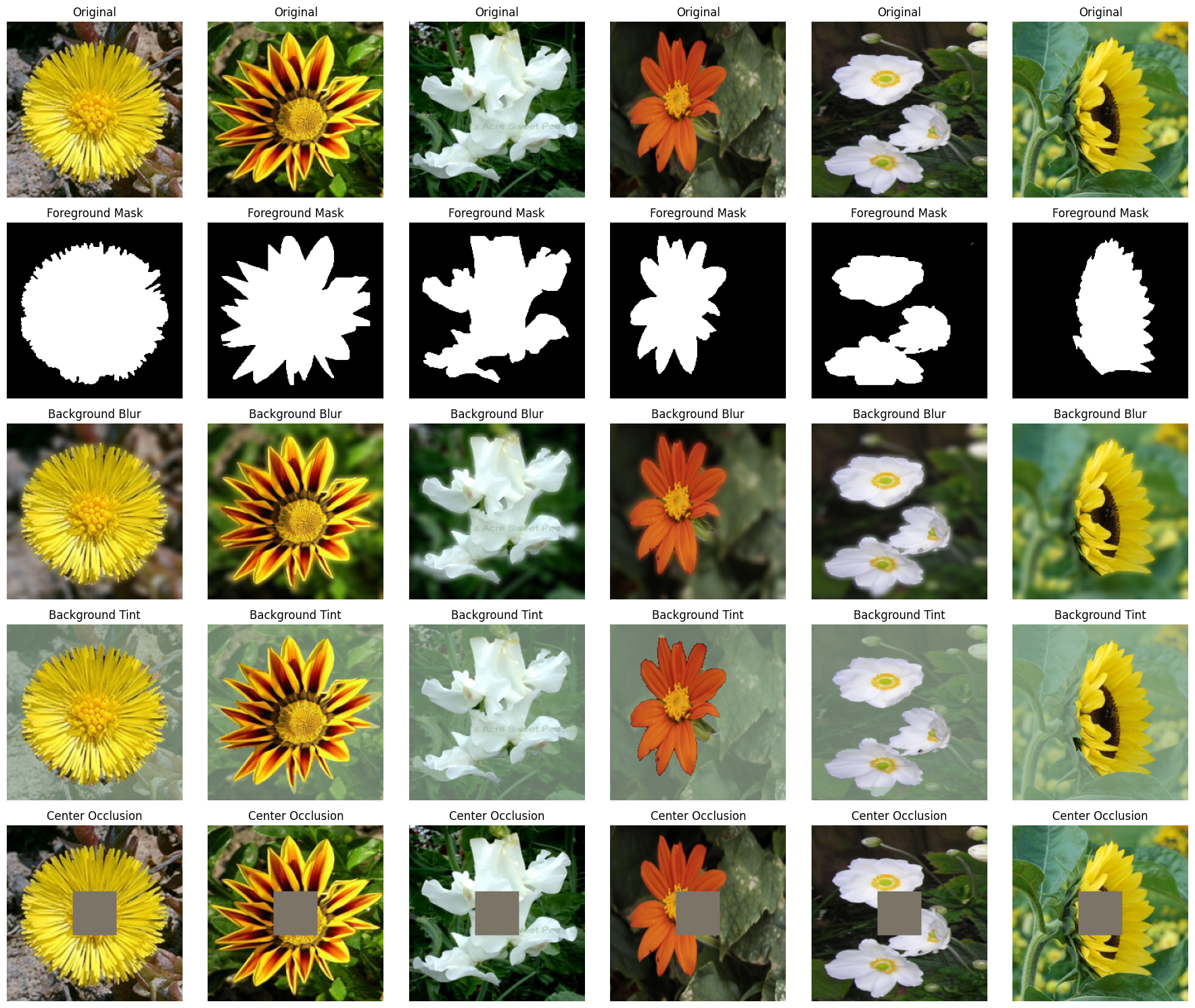}
    \captionof{figure}{Sample counterfactual images.}
    \label{fig:cf-sample}
\end{minipage}
\end{center}

\subsubsection{Learning Objective}
The original I-JEPA loss teaches the model to predict hidden representations. We add a second objective that teaches the model to predict how the representation should change after an intervention.
The total loss becomes:

\begin{equation}
    \mathcal{L}_{\mathrm{total}}
    = \mathcal{L}_{\mathrm{JEPA}}
    + \lambda\mathcal{L}_{\mathrm{CF}}.
    \label{eq:total-loss}
\end{equation}

This allows the model to learn both semantic prediction and intervention-aware representations.
\subsubsection{Evaluation Methods}
We trained three models: a pretrained Vision Transformer (ViT), the original I-JEPA with ViTs as the context and target encoders, and our proposed I-JEPA with counterfactual block. We kept all configurations, including image size, patch size, embedding dimension, learning rate, and dropout, consistent across the models. After training, we froze each encoder and trained a linear classifier to compare model performance~\cite{alain2016understanding}.

\revision{We measure the representation distance between each original image and its counterfactual version. Smaller distances indicate lower sensitivity; however, their interpretation depends on the type of intervention. Low sensitivity to nuisance changes is desirable, whereas a stronger response to semantic changes indicates that the representation retains that information. We therefore evaluate robustness through the combination of semantic sensitivity and nuisance stability. The Selective Intervention Ratio (SIR) summarizes the semantic effect relative to the nuisance effect:}

\begin{equation}
    \mathrm{SIR}
    = \frac{E_{\mathrm{semantic}}}{E_{\mathrm{nuisance}}}.
    \label{eq:sir}
\end{equation}

\revision{For a positive nuisance effect, $\mathrm{SIR}>1$ indicates greater sensitivity to semantic than to nuisance counterfactuals. For the reported mean $L_2$ changes, center occlusion can be compared with background blur and background tint separately. Comparisons across models require matched interventions and consistent representation-distance measurement. If the nuisance effect is zero, the ratio is undefined and the two effects should be reported separately.}

\subsection{Results}

The linear-probing results for the three models are presented in Table~\ref{tab:representation-comparison}. The pretrained ViT-B/16 and the original I-JEPA achieved the same best validation accuracy of 77.55\%. The proposed I-JEPA model with counterfactuals achieved the highest accuracy of 78.14\%, an improvement of approximately 0.59 percentage points.

\begin{table}[htbp]
    \centering
    \caption{Final representation comparison using linear probing.}
    \label{tab:representation-comparison}
    \begin{tabular*}{\textwidth}{@{\extracolsep{\fill}}lc@{}}
        \hline
        \textbf{Model} & \textbf{Best validation accuracy (\%)} \\
        \hline
        Pretrained ViT-B/16 & 77.55 \\
        I-JEPA & 77.55 \\
        \textbf{I-JEPA + CF} & \textbf{78.14} \\
        \hline
    \end{tabular*}
\end{table}

\begin{figure}[htbp]
    \centering
    \includegraphics[width=0.8\textwidth]{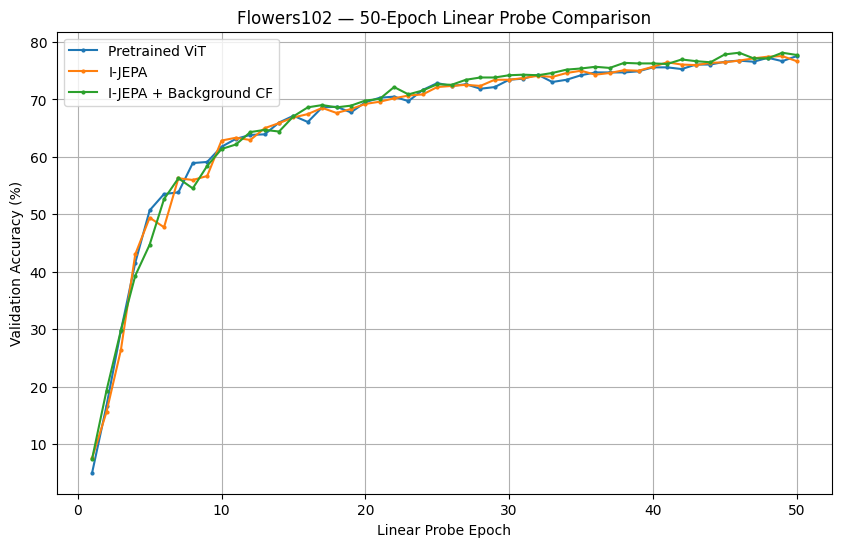}
    \caption{Linear-probing accuracy comparison across the evaluated models.}
    \label{fig:linear-probe-accuracy}
\end{figure}

The linear-probing accuracy comparison is visualized in Figure~\ref{fig:linear-probe-accuracy}.

Table~\ref{tab:intervention-effects} summarizes the effect of each counterfactual intervention on the learned representations. Center occlusion produced the largest mean absolute feature effect, RMS feature effect, and mean $L_2$ representation change. Background blur produced the smallest changes, while background tint had an intermediate effect.

\begin{table}[htbp]
    \centering
    \caption{Counterfactual intervention effect summary. MAE denotes the mean absolute feature effect, and RMS denotes the root-mean-square feature effect.}
    \label{tab:intervention-effects}
    \begin{tabular*}{\textwidth}{@{\extracolsep{\fill}}ccccc@{}}
        \hline
        \textbf{Intervention} & \textbf{MAE} & \textbf{RMS} & \textbf{Mean $L_2$} & \textbf{Std. $L_2$} \\
                              &              &              & \textbf{change}    & \textbf{change} \\[3pt]
        \hline
        Background blur  & 0.08 & 0.11 & 2.83 & 0.74  \\
        Background tint  & 0.10 & 0.13 & 3.48 & 0.98  \\
        Center occlusion & 0.12 & 0.16 & 4.42 & 0.37  \\
        \hline
    \end{tabular*}
\end{table}

We calculated the mean absolute effect (MAE), root-mean-square effect (RMS), mean $L_2$ change, and standard deviation of the $L_2$ change. From the data, we infer
\begin{equation}
    E_{\mathrm{center\ occlusion}}
    > E_{\mathrm{background\ tint}}
    > E_{\mathrm{background\ blur}}.
    \label{eq:intervention-effect-ordering}
\end{equation}

Here, $E_{\mathrm{center}}$ represents the semantic effect $E_{\mathrm{semantic}}$, while $E_{\mathrm{background}}$ and $E_{\mathrm{tint}}$ represent nuisance effects $E_{\mathrm{nuisance}}$. Therefore,
\begin{equation}
    E_{\mathrm{semantic}} > E_{\mathrm{nuisance}}.
    \label{eq:semantic-vs-nuisance}
\end{equation}

\subsubsection{\revision{Robustness Assessment and Limitations}}
\revision{The reported mean $L_2$ changes are 4.42 for center occlusion, 3.48 for background tint, and 2.83 for background blur (Table~\ref{tab:intervention-effects}). The larger response to the semantic counterfactual than to the nuisance counterfactuals is consistent with the selective-sensitivity criterion used in our robustness assessment. However, the nuisance responses remain nonzero, so the results indicate relative selectivity rather than complete nuisance invariance. }

Using the reported mean $L_2$ changes, the SIR is approximately $4.42/2.83 = 1.56$ for center occlusion relative to background blur and $4.42/3.48 = 1.27$ relative to background tint. These are ratios of the reported mean effects, not means of per-image ratios. Intervention effects are not reported for the baseline encoders, so these results do not establish improved selective sensitivity over the baselines. Differences in intervention strength, edited area, and editing artifacts may also contribute to the observed ordering. The reported best validation accuracies do not constitute held-out test results, and the 0.59-percentage-point gain should not be interpreted as statistically significant without repeated-run uncertainty estimates.

\section{Conclusion}
\revision{This paper introduced a counterfactual intervention-aware extension of I-JEPA to study the robustness of learned representations. Our experiments use counterfactuals as controlled tests of whether a representation responds more strongly to semantic changes than to nuisance changes. Under our operational criterion, a more robust representation retains sensitivity to semantic information while remaining relatively stable under nuisance variations. The counterfactual block predicts intervention-induced changes, while representation distances and the Selective Intervention Ratio assess the desirability of the measured response pattern.}

\revision{The reported representation changes are greater for flower-center occlusion than for background blur or tint, consistent with the desired semantic-over-nuisance ordering. This supports relative selectivity for the evaluated interventions, rather than a guarantee of robustness to all visual changes or proof of causal feature discovery. Classification accuracy is complementary to this assessment and does not replace it. Future work will compare intervention sensitivity across baseline models, control intervention strength and edited area, and extend the evaluation to concept-level interventions and additional datasets.}

\begingroup
\sloppy
\hbadness=10000
\bibliographystyle{splncs04-citeorder}
\bibliography{references}
\endgroup

\end{document}